\documentclass[wcp]{jmlr}

\usepackage{longtable}
\usepackage{booktabs}
\usepackage{rotating}
\usepackage{placeins}
\usepackage{nicefrac}

\jmlryear{2014}
\jmlrworkshop{Neural Connectomics Workshop}

\title{Efficient combination of pairwise feature networks}

\author{\Name{Pau Bellot} \Email{pau.bellot@upc.edu}\\
 \addr Department of Signal Theory and Communications, Technical University of Catalonia,
UPC-Campus Nord, C/ Jordi Girona, 1-3, 08034, Barcelona, Spain.
 \AND
 \Name{Patrick E. Meyer} \Email{patrick.meyer@ulg.ac.be}\\
 \addr Bioinformatics and Systems Biology (BioSys),
Faculty of Sciences, Universit\'{e} de Li\`{e}ge (ULg),
27 Blvd du Rectorat, 4000 Li\`{e}ge, Belgium. \url{http://www.biosys.ulg.ac.be/}}

\begin{document}

\maketitle

\begin{abstract}
This paper presents a novel method for the reconstruction of a neural network connectivity using calcium fluorescence data. We introduce a fast unsupervised method to integrate different networks that reconstructs structural connectivity from neuron activity. Our method improves the state-of-the-art reconstruction method General Transfer Entropy (GTE). We are able to better eliminate indirect links, improving therefore the quality of the network via a normalization and ensemble process of GTE and three new informative features. The approach is based on a simple combination of networks, which is remarkably fast. The performance of our approach is benchmarked on simulated time series provided at the connectomics challenge and also submitted at the public competition. \end{abstract}

\section{Introduction}

Understanding the general functioning of the brain and its learning capabilities as well as 
the brain structure and some of its alterations caused by disease, is a key step towards  a treatment of epilepsy, Alzheimer's disease and other neuropathologies.

This could be achieved by recovering neural networks from activities. A neural network is a circuit formed by a group of connected (physically or by means of neural signals) neurons that performs a given functionality. These circuits are responsible for reflexes, senses, as well as more complex processes such as learning and memory.

Thanks to fluorescence imaging, we can easily measure the activity of a group of neurons. The changes of fluorescence recorded from the neural tissue are proved to be directly corresponding to neural activity. With calcium imaging one can study the neural activity of a population of neurons simultaneously allowing to uncover the function of neural networks. 

But, recovering the exact wiring of the brain (connectome) including nearly 100 billion neurons that have on average 7000 synaptic connections to other neurons is still a daunting task. Hence, there is a growing need for fast and accurate methods able to reconstruct these networks.
That is why the ChaLearn non-profit organization has proposed the connectomics challenge.
The goal of the competition is to reconstruct the structure of a neural network from temporal patterns of activities of neurons. 

\section{Typical Methods}

There is a wide variety of reconstruction algorithms that are able to infer the network structure from time series.  Even though one of the least controversial necessary criterion to establish a cause-effect relationship is temporal precedence, many causal inference algorithms only require  conditional independence testing \cite{pearl2000causality}, or, more recently, joint distribution of pairs of variables \cite{Janzing10tellingcause}. The work of Clive Granger has lead to a framework
that has received a lot of attention due to its  simplicity and the successful results \cite{Popescu2013causality}.

\subsection{Correlation with discretization}
Here, we present a quick review of the simplest method to reconstruct a network, based on the correlation coefficient. The correlation is a standard method to quantify the statistical similarity between two random variables $X$ and $Y$  and it is defined as:

\begin{equation}	
corr(X,Y)=\frac{cov(X,Y)}{\sigma_X \sigma_Y}=\frac{E[(X-\mu_X)(Y-\mu_Y)]}{\sigma_X \sigma_Y}
\end{equation}

where  $\mu_X$ and $\mu_Y$ are respectively the expected values of $X$ and $Y$, and $\sigma_X$ and  $\sigma_Y$ are the  standard deviations.

Once we have a correlation coefficient between each pair of neurons, we can construct a
co-activity network. If the correlation is greater than a threshold, then the neurons are connected in an undirected way, this strategy is presented at \cite{butte2000mutual}, where instead of using the correlation they use Mutual Information which can be seen as a non-linear dependency measure.

\subsection{Generalized Transfer Entropy}\label{ssec:GTE}
Here, we present a quick review of one of the state of the art methods, the Transfer Entropy (TE) \cite{schreiber2000measuring} based measure of effective connectivity called Generalized Transfer Entropy, or GTE \cite{Stetter2012}.

The basic idea behind Granger 	causality to test if the  observations of time series of two variables A and B are due to the underlying process ``A causes B'' rather than ``B causes A'', is to fit different predictive models A (present time) and B (present time) as a function of A (past times) and B (past times). If A can be better predicted from past values of A than from past values of B, while B is also better predicted by A, then we have an indication for A being the cause of  B.

Based on this idea, several  methods have been derived in order to improve the results. These methods incorporate the frequency domain analysis instead of a time domain analysis \cite{nolte2010localizing}. One recent idea is to add contemporaneous values of B to predict A and vice versa to take into account instantaneous causal effect, due for instance to insufficient time resolution \cite{moneta2005graph}. 

Therefore, GTE can be seen as a reconstruction algorithm of causal networks based solely on pairwise interactions.

\section{Our proposal: Unsupervised ensemble of CLRed pairwise features}

If $A$ activates $B$ and this last one activates $C$, is very likely that 
co-activity networks will find a strong dependency between  $A$ and  $C$ even though the latter is an indirect 
link. Gene Network Inference methods have proposed different strategies to eliminate these indirect links  \cite{butte2000mutual,margolin2006aracne,meyer2007information}. 

One of these strategies is the Context Likelihood or Relatedness network (CLR) method \cite{faith2007large}. 
In order to do so, the method derives a score that is associated to the 
empirical distribution of the score values. Consider a score $S_{i,j}$ indicating the strength of an alleged connection between two neurons $i$ and $j$. Let us call $\mu_i$ and $\sigma_i$ the mean and standard deviation of this score over all neurons connected to $i$. The asymmetric standardized score is given as:
\begin{equation}
z_i=max \left( 0,\frac{S_{i,j}-\mu_i}{\sigma_i} \right)
\end{equation}

Finally, the symmetrized score is given by: $z_{ij}=\sqrt{z^2_i+z^2_j}$. This method has a complexity of $O(n^2)$,  $n$ being the number of neurons, and requires a symmetric matrix.

Our unsupervised ensemble of pairwise features uses the CLR algorithm to eliminate indirect links and normalize the network before assembling the different CLRed pairwise features. With the second step we are able to eliminate more indirect links that are still present at one reconstructed network but not at the others. This idea comes from \cite{modENCODE,marbach2012wisdom}, where the authors propose an algorithm to 
integrate different network inference methods to construct a community networks which is capable of stabilizing the results and recover a good network.  Their state-of-the-art method to combine networks is based on rank averaging.
The individual ranks of each link are added together to compute the
final rank. Then, the final list is computed sorting these score decreasingly. 
This method is also known as {\emph Ranksum}, and will be referred as  $RS$ in the paper.

Instead of this procedure, our proposal that will be referred as $CLRsum$ or $CS$ is formulated as follows:
\begin{equation}
CS := \sum_{i}^N CLR   \left(feature_i \right)
\end{equation}\label{eq:CLRed}

A description of
the workflow of our network reconstruction process is available in Figure~\ref{fig:CLRsum} in the  Supplementary Material (see section~\ref{sc:SM}).
In this case, we have used four features that are defined in the following subsections.
\subsection{Feature 1: Symmetrized GTE}
The first pairwise measure is a modification of the state-of-the-art method GTE (see section \ref{ssec:GTE}), since we apply the CLR method  the recovered network should be undirected. Indeed, the GTE method 
provides a non-symmetric score ($gte_{i\rightarrow j} \ne gte_{j\rightarrow i}$), we symmetrize it by taking the most conservative score recovered by GTE. This symmetrized GTE network is  denoted as $GTE_{sym}$, and is defined as follows: $gte_{i,j}=min ( gte_{i\rightarrow j},gte_{j\rightarrow i} )$.

\subsection{Feature 2: Correlation of the extrema of the signals} 
The second pairwise measure  is based on the correlation of both signals ($corr(X_i, X_j)$). 
But, doing so we are not able to discriminate between true regulations and indirect effects or light scattering effects. 
We observed experimentally with the provided networks and their respective ground-truth, that the correlation between the 
signals when one of both neurons is spiking is statistically more informative than the plain correlation.

The quantile $q_{k\%}(x)$ is the data value of $x$ where we have $k$ \% of the values of $x$ above it. The higher the quantile the stronger the statistical correlation between the measure and the connectome network. However, in order to be able to recover a non-spurious correlation  at least several hundreds of samples are required.
First, we capture the quantile $\alpha$\% of both signals, and compute the correlation using only the points of both signals that are above the quantile:

\begin{equation}
\begin{aligned}
\text{Let} \; t_k  := X_i(t) \ge q_{\alpha\%}(X_i)  &\;  \text{and}  \; t_l  := X_j(t) \ge q_{\alpha\%}(X_j) \\  
 ct_{i,j}  =corr  \left( X_i (t_n)  ,X_j (t_n)  \right) &\;  \text{with} \;  t_n:= t_k  \cup t_l 
  \end{aligned}
\end{equation}

Computing  previous equation  between each pair of different neurons we obtain the $CT_{\alpha\%}$ network.

\subsection{Feature 3: mean squared error of  difference signal}
The third pairwise feature that we have found experimentally, is complementary to  feature 2. Instead of computing the correlation on the spikes, this feature uses the mean squared error of the points where the two signals disagree the most (once both have been normalized by a centering and scaling). The normalization process is defined as 
$ X^s_i:=\nicefrac{(X_i-\mu_{X_i})}{\sigma_{X_i}}$.

First, we compute the difference between the two scaled different signals ($i \ne j$) and keep the points where they differ the most. To get such particular time points, we also rely on an small quantile $\alpha$\%:
\begin{equation}
\begin{aligned}
\text{Let}\; f_{i,j} (t)=X^s_i (t) - X^s_j (t) \;& \text{and} \;t_k  :=  d_{i,j}(t) \ge q_{\alpha\%}(f_{i,j})  \\
\text{Then} \; X'_i := X^s_i (t_k) & \; X'_j := X^s_j (t_k) 
\end{aligned}
\end{equation}

Once  the points of interest ($t_k$) are extracted, the mean square error between $p_{i,j}:=X'_i-X'_j$
 and $p_{j,i}:=X'_j - X'_i$ is computed. This leads to a non-symmetrical measure. As explained before, CLR requires a symmetric matrix. In order to symmetrize the matrix we  take the minimum of $p^2_{i,j}$ and $p^2_{j,i}$  as has been done in feature 1: $ cd_{i,j}  = \min \left( (X'_i-X'_j)^2,(X'_j - X'_i)^2 \right) $.
 
This measure is computed between each pair of different neurons to obtain the  $MD_{\alpha\%}$  network.

 \subsection{Feature 4: range of difference signal}
 The last pairwise measure that we have found correlated to the connectome is the range of the difference between  two neuron signals.

For every pair of neurons we compute the difference between the two different signals ($i \ne j$):
$df_{i,j}:=X_i - X_j$

Then, the measure captures the range of $df_{i,j}$, but in order to be robust to the presence of noise the range is not the difference between the largest and smallest values of $df_{i,j}$, but the average over the 10th maximal/minimal values of  $df_{i,j}$. 
This measure is computed between each pair of different neurons to obtain the network $RD$. 
In order to obtain a similarity network, 
we invert the network as follows:
 \begin{equation}
RD := max(R)-R \; \text{with}\; diag(RD)=0
 \end{equation}

\section{Experiments}
The performance of our algorithm is benchmarked  on the data provided at the ChaLearn connectomics challenge. The  data reproduces the dynamic behavior of real networks of cultured neurons.  The simulator also  includes the typical real  defects of the calcium fluorescence technology: limited time resolution and light scattering artifacts (the activity of given neuron influences the measurements of nearby neurons) \cite{Stetter2012}.

The challenge provides different datasets that have distinct properties, we will use the datasets where the network structure is also provided, i.e, 10 big datasets of 1000 neurons and 5 small datasets of 100 neurons.
The network inference problem can be seen as a binary decision problem:
after the thresholding of the network provided by the algorithm, the final decision can be seen as a classification: 
for each possible pair of neurons, the algorithm either define a connection or not. Therefore, the performance evaluation can be assessed with  the 
usual metrics  of machine learning like Receiver Operating Characteristic (ROC) and Precision  Recall (PR) curves.  The ChaLearn Connectomics proposes as a global metric the use of the area under the ROC curve (AUC),
however, ROC curves can present an overly optimistic view of an algorithm's performance if there is a large skew in the class distribution, as typically encountered in sparse network problems. To tackle this problem, precision-recall (PR) curves have been proposed as an alternative to ROC curves \cite{sabes1995advances}.
For this reason, we present in Table \ref{tb:AUPR} the Area Under PR curve (AUPR) and 
compare our method with GTE and the state-of-the-art combination of these features {\em Ranksum}.  The results of GTE are obtained with the software available online \cite{dherkova} using as conditioning levels the values $\{0.05, 0.10\}$ for the iNet1-Size100-CC's networks and $\{0.15, 0.20\}$ for the big datasets. We have also computed a statistical test to discard non-significant results. First,  we compute the contribution of each link to the area under the curve and then we apply the Wilcoxon test  on the resulting vectors \cite{hollander2013nonparametric}. If the best result of each dataset have a p-value smaller than $5\%$ it is typed in italic font and boldfaced.

 Table~\ref{tb:AUPR}  shows the performance of our  individual networks ($CT_{0.1\%}$, $MD_{0.1\%}$, $RD$) 
and we can observe that it depends on the properties of the network (high/low-connectivity or high/low-activity). 

\begin{table}[h!]
 \centering
\resizebox{\linewidth}{!}{
 \begin{tabular}{c|*{5}{c}||cc}
  \toprule
   & $I$ & $II$ &$III$ & $IV$ &  $V$  &    \\
  & $GTE$ & $corr$ &$CT_{0.1\%}$ & $RD_{0.1\%}$ &  $MD_{0.1\%}$  &$CS \left(I^*,III,VI,V \right) $ 
  & $RS\left(I,III,VI,V \right)$ \\ 
   \midrule
highcc & 0.163 & 0.051 & 0.088 & 0.166 & 0.125 & \textbf{\emph{0.330}} & 0.184 \\ 
  highcon & 0.199 & 0.030 & 0.120 & 0.125 & 0.073 & \textbf{\emph{0.241}} & 0.184 \\ 
  iNet1-Size100-CC01inh & \textbf{\emph{0.242 }}& 0.117 & 0.117 & 0.106 & 0.123 & 0.180 & 0.158 \\ 
  iNet1-Size100-CC02inh & \textbf{\emph{0.247}} & 0.113 & 0.150 & 0.103 & 0.120 & 0.223 & 0.181 \\ 
  iNet1-Size100-CC03inh & \textbf{\emph{0.333}} & 0.116 & 0.198 & 0.130 & 0.131 & 0.314 & 0.237 \\ 
  iNet1-Size100-CC04inh &\textbf{\emph{0.398}} & 0.120 & 0.206 & 0.187 & 0.158 & 0.394 & 0.297 \\ 
  iNet1-Size100-CC05inh & 0.366 & 0.120 & 0.208 & 0.288 & 0.179 & \textbf{\emph{0.423}} & 0.366 \\ 
  iNet1-Size100-CC06inh & 0.538 & 0.204 & 0.188 & 0.371 & 0.318 & \textbf{\emph{0.582}} & 0.480 \\ 
  lowcc & 0.085 & 0.015 & 0.085 & 0.031 & 0.022 & \textbf{\emph{0.126}} & 0.083 \\ 
  lowcon & 0.093 & 0.023 & 0.025 & 0.024 & 0.031 & \textbf{\emph{0.196}} & 0.125 \\ 
  normal-1 & 0.164 & 0.028 & 0.085 & 0.110 & 0.061 & \textbf{\emph{0.251}} & 0.155 \\ 
  normal-2 & 0.169 & 0.028 & 0.105 & 0.095 & 0.048 &\textbf{\emph{0.242}} & 0.153 \\ 
  normal-3-highrate & 0.201 & 0.057 & 0.037 & 0.143 & 0.073 & \textbf{\emph{0.293}} & 0.181 \\ 
  normal-3 & 0.193 & 0.025 & 0.098 & 0.094 & 0.044 & \textbf{\emph{0.260}} & 0.150 \\ 
  normal-4-lownoise & 0.141 & 0.030 & 0.120 & 0.086 & 0.053 & \textbf{\emph{0.271}} & 0.156 \\ 
  normal-4 & 0.139 & 0.026 & 0.085 & 0.082 & 0.046 & \textbf{\emph{0.254}} & 0.140 \\ 
   \midrule
    Mean  & 0.229 & 0.069 & 0.120 & 0.134 & 0.100 & \textbf{0.286} & 0.202 \\ 
\bottomrule
\end{tabular}
}
\caption{Area Under Precision Recall scores for each inference method at the datasets of the connectomics challenge (the higher the better). The {\emph Ranksum} makes use of the original pairwise inferred networks while our method use the symmetrized GTE (denoted as $I^*$). The best statistically significant results tested with a Wilcoxon test are highlighted.}\label{tb:AUPR}
\end{table}
We also compare our community based approach with the state-of-the-art {\emph Ranksum} approach, which is shown at the last column. 
Note that the {\emph Ranksum} makes use of the original pairwise derived networks and our method used the symmetrized GTE (denoted as $I^*$).
We can observe that our normalization and simple combination is able to improve the quality of the individual recovered networks and also improves the state-of-the-art community {\emph Ranksum}.  As shown in the table, our approach is competitive  even though our method does not recover a directed network as GTE does.  It is worth noting that using AUC as a metric we obtain similar conclusions. Table~\ref{tb:AUC} with AUC results  is available in the Supplementary Material (see section~\ref{sc:SM}).

Additionally to the results shown at table~\ref{tb:AUPR}  we also have used our method in the test and validation networks where the network is unknown. Using the connectomics submission tool we obtain 0.90402 score of Area under the ROC curve, and we would have been ranked in the 30th position.  

As stated previously the big advantage of our method is the low complexity $O(n^2)$. The CPU time needed to compute the different features for the big datasets in a 2 x Intel Xeon E5 2670 8C (2.6 GHz), has a mean of 1282.86 minutes for the GTE, 3.31 minutes for $CT_{0.1\%}$,  66.53 minutes for $MD_{0.1\%}$ and 21.17 minutes for $RD$. The process of CLRsum is almost instantaneous once we have the individual features, and therefore the  computation time is the sum of the time needed to compute the individual features. Hence, our proposal improves  GTE with a negligible overload of time. 

\section{Conclusion}
An unsupervised network inference method for neural connectomics has been presented. 
This method improves the state-of-the-art network inference method GTE relying on $CLRsum$ consensus among GTE and  three  new informative features. 

We have compared our method experimentally to two state-of-the-art network inference methods, namely GTE and correlation network, on the connectomics challenge datasets. The experimental results showed that
our proposal is competitive with  state-of-the-art algorithms.
\acks{This work has been partially supported by the Spanish ``MECD'' FPU  Research Fellowship,
the  Spanish ``MICINN'' project TEC2013-43935-R and the Cellex foundation.}

\section{Supplemental Material}\label{sc:SM}
\begin{figure}[htb]  
\centering
\includegraphics[width=\textwidth]{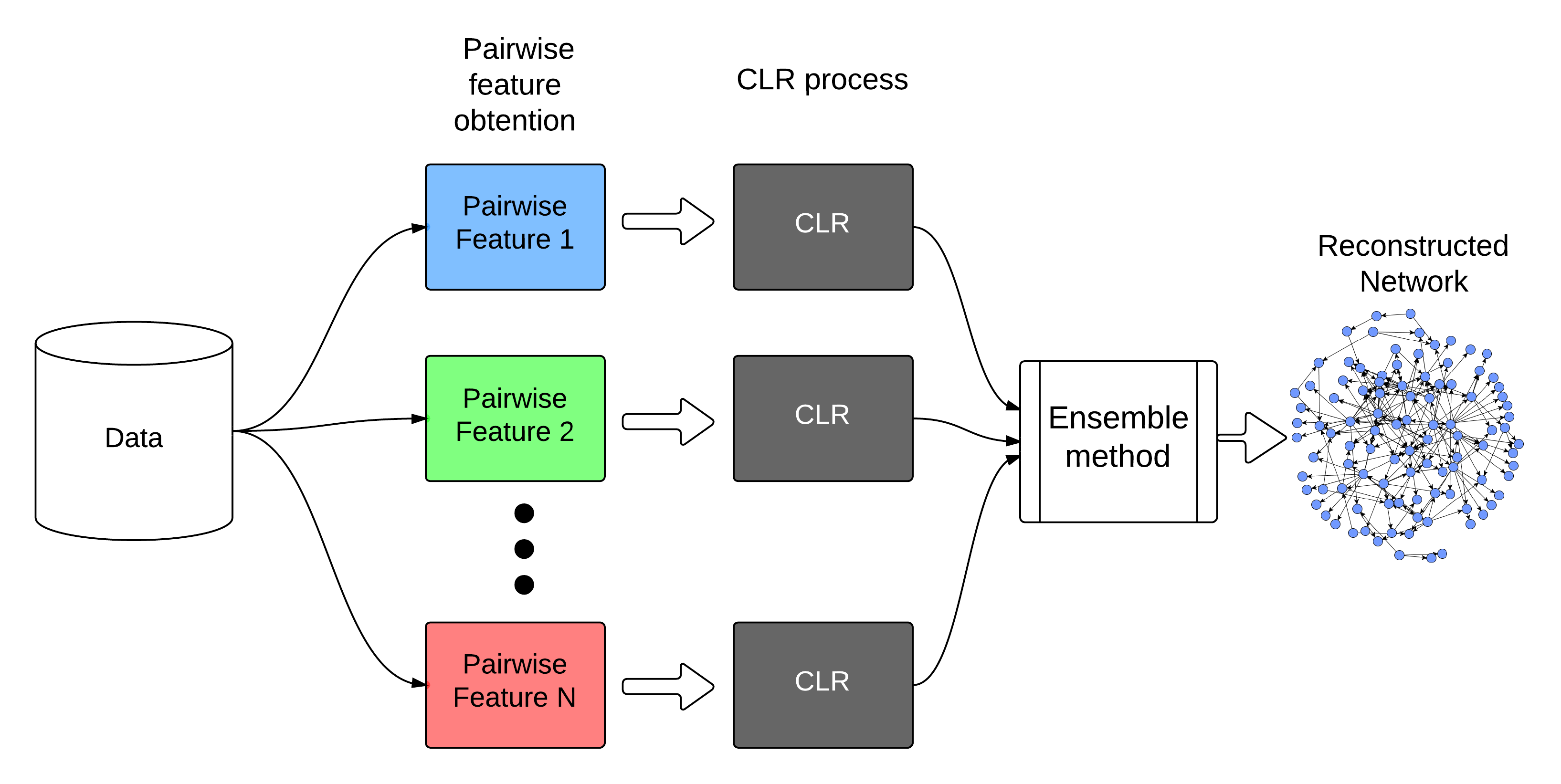}
\caption{Workflow of the network reconstruction process.}\label{fig:CLRsum}
\end{figure}

\begin{table}[htb]
 \centering
\resizebox{\textwidth}{!}{
 \begin{tabular}{c|ccccc||cc}
  \toprule
   & $I$ & $II$ &$III$ & $IV$ &  $V$  &    \\
  & $GTE$ & $corr$ &$CT_{0.1\%}$ & $RD_{0.1\%}$ &  $MD_{0.1\%}$  &$CS \left(I^*,III,VI,V \right) $ 
  & $RS\left(I,III,VI,V \right)$ \\ 
   \midrule
  highcc & \textbf{\emph{0.898}} & 0.755 & 0.791 & 0.871 & 0.863 & 0.891 & 0.885 \\ 
  highcon & 0.898 & 0.595 & 0.826 & 0.806 & 0.776 &  \textbf{\emph{0.920 }} & 0.896 \\ 
  iNet1-Size100-CC01inh &  \textbf{\emph{0.705}} & 0.556 & 0.559 & 0.513 & 0.570 & 0.670 & 0.640 \\ 
  iNet1-Size100-CC02inh & 0.703 & 0.563 & 0.653 & 0.525 & 0.576 &  \textbf{\emph{0.733}} & 0.684 \\ 
  iNet1-Size100-CC03inh & 0.761 & 0.582 & 0.726 & 0.591 & 0.616 &  \textbf{\emph{0.814}} & 0.753 \\ 
  iNet1-Size100-CC04inh & 0.789 & 0.577 & 0.731 & 0.662 & 0.646 &  \textbf{\emph{0.848}} & 0.787 \\ 
  iNet1-Size100-CC05inh & 0.793 & 0.568 & 0.697 & 0.776 & 0.679 &  \textbf{\emph{0.875}} & 0.818 \\ 
  iNet1-Size100-CC06inh & 0.869 & 0.753 & 0.666 & 0.822 & 0.816 &  \textbf{\emph{0.919}} & 0.884 \\ 
  lowcc & 0.864 & 0.571 & 0.818 & 0.680 & 0.658 &  \textbf{\emph{0.883}} & 0.839 \\ 
  lowcon & 0.733 & 0.691 & 0.683 & 0.689 & 0.690 & \textbf{\emph{0.829}}& 0.778 \\ 
  normal-1 &  \textbf{\emph{0.891}} & 0.681 & 0.801 & 0.837 & 0.799 & 0.888 & 0.874 \\ 
  normal-2 &  \textbf{\emph{0.891}} & 0.699 & 0.830 & 0.826 & 0.788 & 0.887 & 0.877 \\ 
  normal-3-highrate &  \textbf{\emph{0.888}} & 0.785 & 0.700 & 0.847 & 0.812 &  0.879 & 0.874 \\ 
  normal-3 & 0.883 & 0.683 & 0.811 & 0.813 & 0.774 &  \textbf{\emph{0.886}} & 0.872 \\ 
  normal-4-lownoise & 0.884 & 0.708 & 0.808 & 0.803 & 0.774 &  \textbf{\emph{0.888}} & 0.866 \\ 
  normal-4 & 0.879 & 0.681 & 0.796 & 0.793 & 0.763 &  \textbf{\emph{0.885}} & 0.861 \\ 
  \midrule
  Mean & 0.833 & 0.653 & 0.743 & 0.741 & 0.725 & \textbf{0.856} & 0.824 \\
    \bottomrule
\end{tabular}
}
\caption{AUC scores for each inference method at the different datasets of the connectomics challenge. The best methods are typed in boldface.}\label{tb:AUC}
\end{table}
\end{document}